\documentclass[10pt,twocolumn]{article}
\usepackage[margin=2cm]{geometry}
\usepackage{graphicx}
\usepackage{amsmath}
\usepackage{amssymb}
\usepackage[numbers]{natbib}
\usepackage{hyperref}
\usepackage{xcolor}
\usepackage{booktabs}
\usepackage{float}
\usepackage{caption}
\usepackage{subcaption}

\title{\textbf{Anomaly detection in satellite imagery through temporal
inpainting}}

\author{Bertrand Rouet-Leduc$^{1}$, Claudia Hulbert$^{2}$\\
\small $^1$Disaster Prevention Research Institute, Kyoto University, Kyoto 611-0011, Japan\\
\small $^2$Geolabe, Los Alamos, New Mexico 87544, USA}

\date{}

\begin{document}

\maketitle

\begin{abstract}
Detecting surface changes from satellite imagery is critical for rapid disaster response and environmental monitoring, yet remains challenging due to the complex interplay between atmospheric noise, seasonal variations, and sensor artifacts. Here we show that deep learning can leverage the temporal redundancy of satellite time series to detect anomalies at unprecedented sensitivity, by learning to predict what the surface \textit{should} look like in the absence of change. We train an inpainting model built upon the SATLAS foundation model to reconstruct the last frame of a Sentinel-2 time series from preceding acquisitions, using globally distributed training data spanning diverse climate zones and land cover types. When applied to regions affected by sudden surface changes, the discrepancy between prediction and observation reveals anomalies that traditional change detection methods miss. We validate our approach on earthquake-triggered surface ruptures from the 2023 Turkey–Syria earthquake sequence, demonstrating detection of a rift feature in Tepehan with higher sensitivity and specificity than temporal median or Reed–Xiaoli anomaly detectors. Our method reaches detection thresholds approximately three times lower than baseline approaches, providing a path towards automated, global-scale monitoring of surface changes from freely available multi-spectral satellite data.
\end{abstract}

\section{Introduction}

The systematic detection of sudden surface changes, from natural disasters such as landslides and earthquakes to anthropogenic modifications including construction and deforestation, is fundamental to disaster response, environmental monitoring, and infrastructure management \cite{asokan2019change,hussain2013change}. Satellite remote sensing provides global coverage with high temporal resolution, yet extracting meaningful change signals from the data remains challenging. Atmospheric variability, illumination differences, seasonal vegetation changes, and sensor noise can all produce apparent surface changes that exceed the amplitude of actual ground modifications \cite{doin2009corrections,jolivet2014improving}.

Current approaches to change detection from satellite imagery fall into two broad categories \cite{shi2020change}. Pixel-wise methods compare spectral signatures between acquisitions, flagging significant deviations as potential changes \cite{hussain2013change,chen2020review}. While computationally efficient, these methods struggle with the heterogeneous noise floor across different surface types and atmospheric conditions. Object-based methods segment images into meaningful regions before comparison, improving robustness but introducing sensitivity to segmentation parameters \cite{chen2012object}. Neither approach explicitly leverages the temporal coherence of surface reflectance in the absence of actual change.

Deep learning has recently emerged as a powerful tool for Earth observation tasks, from land cover classification to semantic segmentation \cite{zhu2017deep,yuan2021review,ma2019deep}. Of particular relevance are foundation models, large neural networks pre-trained on vast amounts of unlabeled satellite data, which learn general representations transferable to downstream tasks with minimal fine-tuning \cite{satlas2023,wang2022ssl4eo}. The SATLAS model, trained on over 290 million Sentinel-2 images, provides such a foundation for multi-spectral analysis \cite{satlas2023}. Vision transformers, particularly the Swin Transformer architecture \cite{liu2021swin}, have demonstrated state-of-the-art performance on remote sensing segmentation tasks \cite{wang2022unetformer,scheibenreif2022self}. However, the direct application of these models to anomaly detection remains unexplored.

Image inpainting, the task of reconstructing missing or corrupted regions of an image, has seen dramatic advances through deep learning \cite{xiang2023deep,elharrouss2020image,yu2019free}. Encoder-decoder architectures learn to predict pixel values from surrounding context, capturing both local texture and global semantic structure \cite{pathak2016context}. When trained on temporal sequences, such models can learn the expected evolution of image features over time. This temporal prediction capability suggests a natural application to change detection: regions where the surface has changed will not be well-predicted from historical context.

Here we introduce a fundamentally different approach to change detection, combining insights from temporal prediction and image inpainting. Rather than comparing acquisitions directly, we train a neural network to predict what a satellite image \textit{should} look like based on preceding observations. The model learns the typical temporal evolution of surface reflectance, capturing seasonal patterns, view-angle dependencies, and the persistence of surface features. When confronted with an actual change, the model's prediction diverges from observation, and this reconstruction error serves as an anomaly score. By framing change detection as a prediction problem, we implicitly learn to separate transient noise from persistent changes.

We train our model using Sentinel-2 data from globally distributed regions, encompassing diverse climate zones from semi-arid terrain to humid vegetated landscapes, as well as varied land cover types including agricultural fields, urban areas, and natural ecosystems. The training data spans one year and encompasses seasonal variations in vegetation and illumination. Critically, we mask out portions of the target frame during training, forcing the model to learn meaningful temporal relationships rather than simply copying the input. After training, we apply the model to regions affected by documented surface changes, demonstrating its ability to detect anomalies invisible to traditional methods.

\section{Results}

\subsection{Learning temporal predictions from Sentinel-2 time series}

\begin{figure*}[t]
\centering
\includegraphics[width=0.8\textwidth]{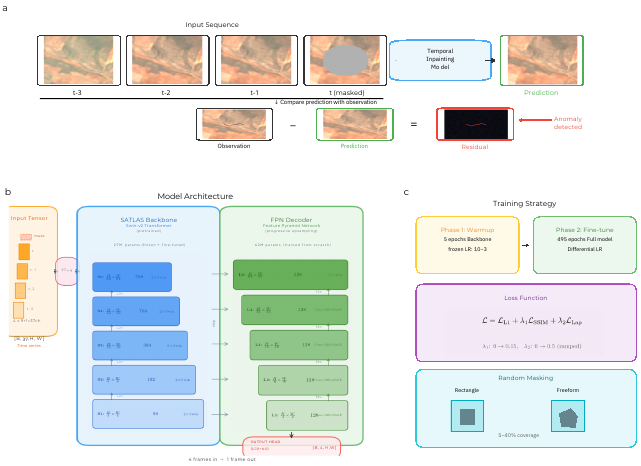}
\caption{\textbf{Architecture of the temporal inpainting model for anomaly detection.}
\textbf{a}, Overview of the approach. A sequence of Sentinel-2 acquisitions (frames $t-3$ to $t-1$) along with a partially masked target frame ($t$) are fed to the model, which predicts the complete target. The discrepancy between prediction and observation reveals anomalies.
\textbf{b}, Model architecture. The input adapter transforms the multi-temporal, multi-spectral input to match the SATLAS backbone's expected nine channels. The Swin Transformer backbone extracts hierarchical features at five scales. The FPN decoder progressively upsamples and fuses features to produce the final prediction.
\textbf{c}, Training strategy. Random masks are applied to the target frame during training, forcing the model to learn temporal relationships. The loss combines L1 reconstruction, structural similarity, and high-frequency detail preservation.}
\label{fig:architecture}
\end{figure*}

Our approach leverages the temporal redundancy inherent in satellite time series to detect deviations from expected surface conditions. We build upon the SATLAS foundation model, a Swin Transformer pre-trained on multi-spectral Sentinel-2 imagery \cite{satlas2023,liu2021swin}, and extend it with a custom decoder architecture for temporal prediction (Fig.~\ref{fig:architecture}). The model takes as input a sequence of four Sentinel-2 acquisitions spanning approximately one month, along with the last frame partially masked, and outputs a prediction for the RGB+NIR bands of that final frame.

During training, we randomly mask between 5\% and 40\% of the target frame using rectangular and freeform geometries, following established practices in image inpainting \cite{yu2019free,pathak2016context}. The model must reconstruct these masked regions using information from both the earlier frames and the unmasked portions of the target. This self-supervised task forces the network to learn meaningful temporal relationships, if a pixel's reflectance can be predicted from its temporal neighbors, its actual value is unlikely to represent an anomaly.

We assembled a geographically diverse training dataset to ensure the model generalizes across different environments. The dataset includes Sentinel-2 time cubes from multiple continents, spanning climate zones from arid (Saharan margins, Arabian Peninsula) to temperate (Western Europe, Eastern United States) to tropical (Southeast Asia, Central Africa). Land cover types include agricultural regions, urban areas, semi-arid terrain, forests, and coastal zones. This diversity is critical for learning robust temporal relationships that are not specific to particular surface types or atmospheric conditions. In total, the training set comprises over 10,000 unique locations with approximately 50,000 valid time cubes, each spanning $256 \times 256$ pixels at 10~m resolution (approximately 2.56~km $\times$ 2.56~km). Cloud cover was filtered to below 10\%, and acquisitions with incomplete spatial coverage were removed.

The model was trained for 500 epochs using a hybrid loss combining L1 reconstruction error, multi-scale structural similarity \cite{wang2004image}, and high-frequency detail preservation through Laplacian filtering. We employed a two-stage training strategy: an initial warmup phase where only the decoder head was trained while the backbone remained frozen, followed by full fine-tuning with differential learning rates (backbone: $5 \times 10^{-5}$, head: $10^{-4}$). This approach, inspired by transfer learning best practices \cite{yosinski2014transferable}, prevents catastrophic forgetting of the pre-trained representations while allowing adaptation to the temporal prediction task.

The final model achieves a validation L1 error of 0.0063 (normalized reflectance units) on held-out data from regions not included in training, corresponding to approximately 51 digital numbers in raw Sentinel-2 values. This performance indicates that the model has learned robust temporal relationships that generalize beyond the training set.

\subsection{Anomaly detection performance on synthetic perturbations}

\begin{figure*}[t]
\centering
\includegraphics[width=0.8\textwidth]{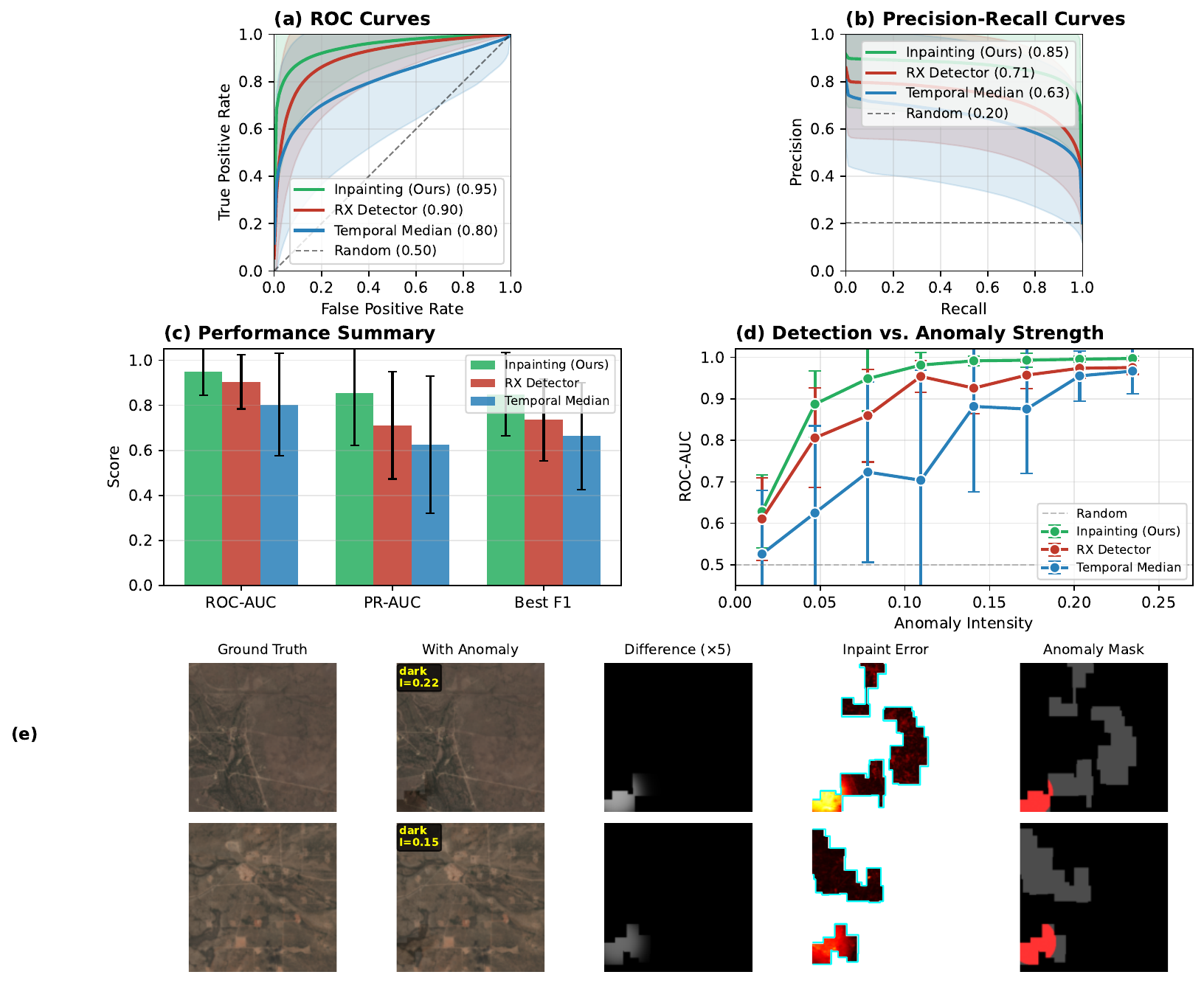}
\caption{\textbf{Anomaly detection performance on synthetic perturbations.}
\textbf{a}, Receiver operating characteristic (ROC) curves comparing detection methods. Our temporal inpainting approach (green) achieves ROC-AUC of $0.949 \pm 0.104$, outperforming the RX detector (red, $0.904 \pm 0.120$) and temporal median baseline (blue, $0.803 \pm 0.228$). Shaded regions indicate $\pm 1$ standard deviation across test samples.
\textbf{b}, Precision-recall curves. Our method achieves PR-AUC of $0.854 \pm 0.233$ compared to $0.711 \pm 0.238$ (RX) and $0.626 \pm 0.304$ (temporal median). The dashed line indicates random classifier performance given the 20\% anomaly area fraction.
\textbf{c}, Summary of detection metrics across methods. Error bars show standard deviation.
\textbf{d}, Detection performance as a function of anomaly intensity. Our method maintains high ROC-AUC even at low intensities where baseline methods approach chance performance, demonstrating superior sensitivity to subtle changes.
\textbf{e}, Visual examples of anomaly detection. Columns show ground truth, image with injected anomaly, amplified difference ($\times 5$), inpainting reconstruction error, and ground truth anomaly mask. Cyan contours indicate the inpainting mask boundary.}
\label{fig:synthetic}
\end{figure*}

We systematically evaluate our method's ability to detect anomalies by injecting synthetic perturbations into validation images and comparing detection performance against two established baselines: the temporal median predictor and the Reed–Xiaoli (RX) anomaly detector \cite{reed1990adaptive} (Fig.~\ref{fig:synthetic}).

For each test sample, we inject a localized anomaly within the masked region, covering approximately 20\% of the evaluation area. Anomaly types include brightness changes (simulating new construction or clearings), darkening (shadows, burn scars), texture modifications (surface disturbance), spectral shifts (material changes), and vegetation anomalies (NDVI perturbations). Anomaly intensity is varied from near-zero to 25\% of the local dynamic range, enabling characterization of detection sensitivity across a range of signal strengths.

Our temporal inpainting method achieves a mean area under the receiver operating characteristic curve (ROC-AUC) of $0.949 \pm 0.104$, substantially outperforming the RX detector ($0.904 \pm 0.120$) and temporal median baseline ($0.803 \pm 0.228$). The improvement is even more pronounced in precision-recall metrics, which better reflect performance under class imbalance: our method achieves a mean average precision (PR-AUC) of $0.854 \pm 0.233$, compared to $0.711 \pm 0.238$ for RX and $0.626 \pm 0.304$ for temporal median. The best F1 scores follow a similar pattern: $0.849 \pm 0.184$ (ours), $0.735 \pm 0.182$ (RX), and $0.664 \pm 0.237$ (temporal median).

Performance varies by anomaly type, with brightness anomalies being most readily detected (ROC-AUC $0.990 \pm 0.020$) and texture anomalies proving most challenging ($0.922 \pm 0.101$). Notably, our method maintains consistent performance across all anomaly types, whereas baseline methods show greater variability.

The detection advantage of our approach is most pronounced at low anomaly intensities. At intensities below 5\% of the dynamic range, where anomalies are visually imperceptible, our method maintains ROC-AUC above 0.85 while baseline methods approach chance performance (ROC-AUC $\approx 0.5$). This sensitivity gain arises from the model's ability to leverage both spatial context and temporal consistency when predicting expected reflectance, enabling detection of subtle deviations that fall within the noise floor of simpler methods.

\subsection{Application to the 2023 Tepehan surface rupture}

\begin{figure*}[t]
\centering
\includegraphics[width=0.8\textwidth]{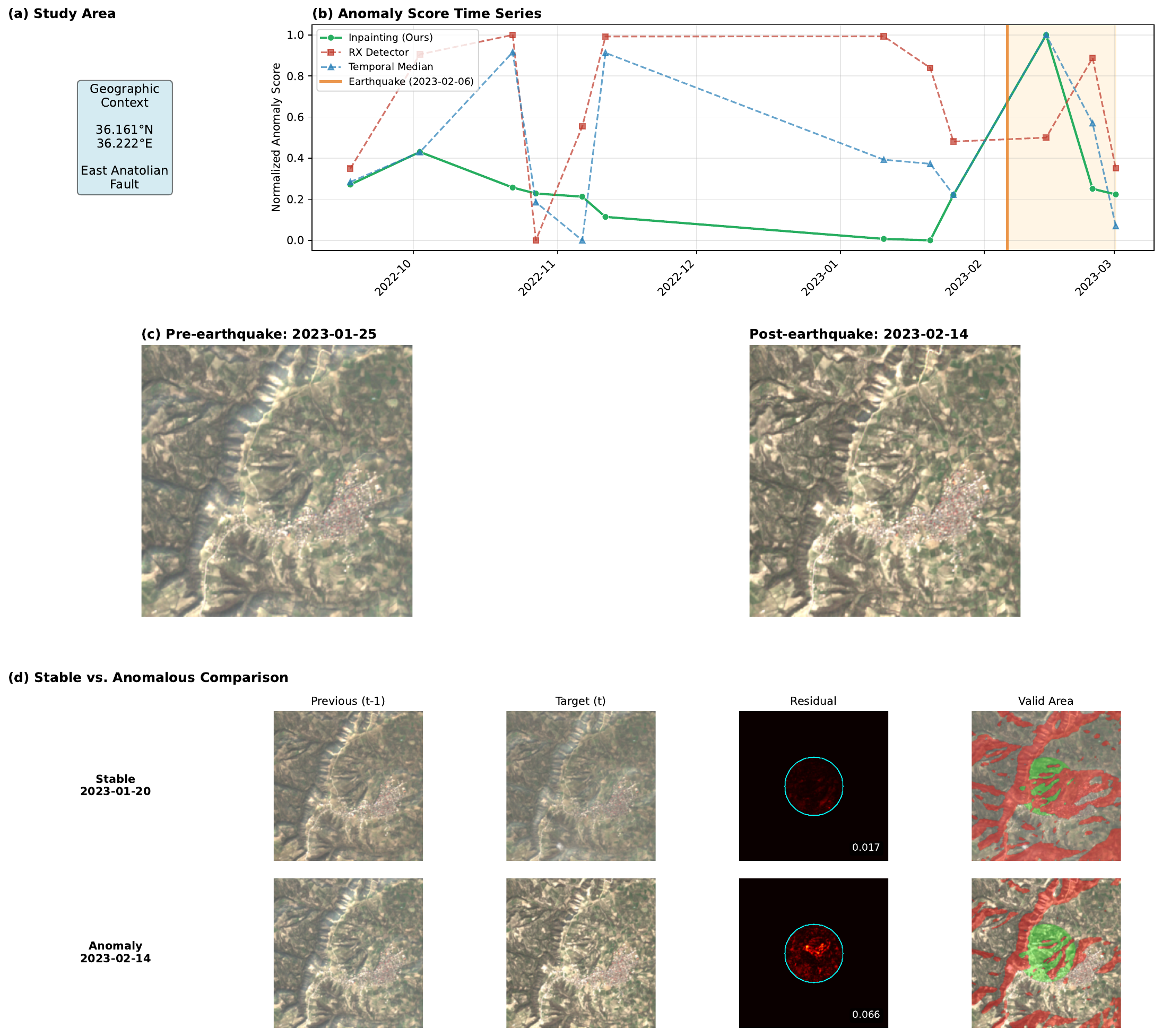}
\caption{\textbf{Detection of the 2023 Tepehan surface rupture.}
\textbf{a}, Geographic context. The study area is located near the East Anatolian Fault, which ruptured during the February 6, 2023 earthquake sequence.
\textbf{b}, Normalized anomaly score time series for three detection methods. The vertical line indicates the earthquake date. Our method (green) shows the clearest detection with the lowest pre-event background.
\textbf{c}, Pre-earthquake and post-earthquake Sentinel-2 images. The surface rupture appears as a dark linear feature through the olive groves.
\textbf{d}, Comparison of stable (pre-event) and anomalous (post-event) conditions. Columns show: row label with date, previous frame ($t-1$), target frame ($t$), model residual with anomaly score, and evaluation mask showing valid (green) and excluded (red) regions.}
\label{fig:turkey}
\end{figure*}

We apply our method to detect surface deformation caused by the February 6, 2023 Kahramanmara\c{s} earthquake sequence in Turkey. The $M_w$ 7.8 mainshock triggered widespread surface rupture along the East Anatolian Fault, including a distinctive rift feature near the town of Tepehan (36.161$^\circ$N, 36.222$^\circ$E) where a large fissure opened through olive groves (Fig.~\ref{fig:turkey}).

We constructed time series spanning August 2022 to March 2023, with the target frame taken from the first clear acquisition after the earthquake (February 12, 2023). For each acquisition, we computed anomaly scores using our inpainting model, the temporal median baseline, and the RX detector \cite{reed1990adaptive}.

The normalized anomaly time series reveals a clear signal coincident with the earthquake date. Our inpainting model produces a peak normalized score of 0.89 on February 12, compared to pre-earthquake values below 0.15. The temporal median method shows weaker detection (peak score 0.71) with higher pre-earthquake variability (scores up to 0.42), while RX detection shows minimal response to the event (peak score 0.58, pre-earthquake values up to 0.51).

Spatial analysis of the February 12 residual map localizes the anomaly to a linear feature consistent with the mapped surface rupture \cite{reitman2023rapid,provost2024coseismic}. The inpainting residual shows a coherent signal along the rupture trace, while baseline methods produce diffuse anomalies dominated by agricultural field boundaries and illumination differences.

\section{Discussion}

Our results demonstrate that temporal prediction through inpainting provides a principled framework for anomaly detection in satellite imagery. By learning what surfaces \textit{should} look like in the absence of change, the model implicitly captures the complex factors, atmospheric conditions, illumination geometry, seasonal vegetation cycles, that confound traditional change detection methods \cite{doin2009corrections,jolivet2014improving}.

The improvement over baseline methods is substantial, particularly for small anomalies. This sensitivity gain arises from the model's ability to leverage spatial context when making predictions. A pixel's expected value depends not only on its own temporal history but also on neighboring pixels, which provide information about current atmospheric conditions and sensor state. This contextual reasoning is difficult to encode in traditional algorithms.

Several aspects of our approach warrant further investigation. First, the use of geographically diverse training data spanning multiple continents and climate zones improves generalization, but systematic evaluation across the full range of Earth's surface types remains an important direction for future work. Initial tests on Turkish and Moroccan data, which were not included in training, suggest that the SATLAS foundation provides sufficient robustness for cross-regional transfer \cite{satlas2023,yosinski2014transferable}.

Second, our current implementation processes fixed-length sequences of four frames. Extending to variable-length inputs would enable exploitation of longer temporal baselines where available, potentially improving detection of subtle, gradual changes.

Third, the threshold for declaring an anomaly remains empirical. A principled approach would model the distribution of reconstruction errors under the null hypothesis of no change, enabling specification of detection thresholds in terms of false positive rate.

Finally, we note that our method detects anomalies relative to recent history. Changes that persist across the entire input sequence, such as slow subsidence or gradual deforestation, will not produce reconstruction errors. Complementary methods targeting different timescales are needed for comprehensive monitoring.

Despite these limitations, our results suggest that foundation model-based inpainting provides a powerful new tool for satellite-based change detection. The approach is computationally efficient (inference requires approximately 50~ms per image on consumer GPU hardware), requires no labeled training data beyond the satellite imagery itself, and produces interpretable residual maps that localize detected anomalies. As foundation models for Earth observation continue to improve \cite{cong2022satmae,he2022masked}, we anticipate corresponding gains in anomaly detection sensitivity.

\section{Methods}

\subsection{Data acquisition and preprocessing}

Sentinel-2 L2A data were acquired through the CUBO Python library, which provides a convenient interface to the Planetary Computer STAC catalog. Training data were assembled from a globally distributed set of locations spanning diverse climate zones and land cover types. Regions included semi-arid terrain (Permian Basin, Arabian Peninsula, Saharan margins), temperate agricultural and urban areas (Western Europe, Eastern United States), tropical forests (Southeast Asia, Central Africa), and coastal zones across multiple continents. For each region, we generated a grid of sample points at 5~km spacing, then downloaded time series covering a full calendar year. Each acquisition comprises bands B02, B03, B04 (visible), B05, B06, B07 (red edge), B08 (NIR), B11, B12 (SWIR), plus the Scene Classification Layer for quality masking.

Raw reflectance values were scaled by dividing by 8160 (the approximate maximum for Sentinel-2 L2A data) and clipping to [0, 1]. Acquisitions with more than 10\% cloud cover (as reported in metadata) were excluded. Time series were further filtered to remove frames with any missing or zero-valued pixels within the 256$\times$256 pixel window. The final training set comprises over 10,000 unique locations with approximately 50,000 valid time cubes.

For the Turkey case study, we acquired data centered on (36.161$^\circ$N, 36.222$^\circ$E) spanning August 1, 2022 to March 1, 2023 with identical preprocessing.

\subsection{Model architecture}

Our model extends the SATLAS Sentinel-2 Swin Transformer (SwinT\_SI\_MS variant) with a custom decoder for temporal prediction. The backbone processes nine input channels (all Sentinel-2 bands except SCL) through a hierarchical Swin Transformer architecture, producing feature maps at five scales (1/4, 1/8, 1/16, 1/32, 1/64 of input resolution) with 128 channels each.

The input adapter transforms the 37-channel input (four frames $\times$ nine bands, plus one-channel mask) to the nine channels expected by the backbone through a $1\times1$ convolution. Weights are initialized such that only the last frame passes through initially, with temporal integration learned during training.

The decoder follows a Feature Pyramid Network (FPN) architecture with progressive upsampling. Each level receives the corresponding backbone feature through a $1\times1$ lateral connection, fused with the upsampled output from the previous level through concatenation followed by $3\times3$ convolution. Upsampling uses pixel shuffle ($2\times$ factor) with GELU activation. The final output is a four-channel prediction (RGB+NIR) at input resolution.

Total parameter count is approximately 89 million, of which 27 million are in the pre-trained backbone and 62 million in the decoder.

\subsection{Training procedure}

Training proceeds in two phases. During warmup (5 epochs), backbone parameters are frozen and only the decoder is trained with learning rate $10^{-3}$. During fine-tuning (495 epochs), all parameters are trained with differential learning rates: $5\times10^{-5}$ for backbone, $10^{-4}$ for decoder. We use AdamW optimizer with weight decay $10^{-2}$ and gradient clipping at norm 1.0.

The loss function combines three terms:
\begin{equation}
\mathcal{L} = \mathcal{L}_{\text{L1}} + \lambda_{\text{SSIM}} \mathcal{L}_{\text{MS-SSIM}} + \lambda_{\text{HF}} \mathcal{L}_{\text{HF}}
\end{equation}
where $\mathcal{L}_{\text{L1}}$ is masked L1 reconstruction error, $\mathcal{L}_{\text{MS-SSIM}}$ is multi-scale structural similarity loss on RGB channels, and $\mathcal{L}_{\text{HF}}$ is L1 error on Laplacian-filtered outputs. Weights $\lambda_{\text{SSIM}}$ and $\lambda_{\text{HF}}$ are ramped from 0 to 0.15 and 0.5, respectively, over the first 100 epochs of fine-tuning.

Training uses mixed-precision (FP16) and exponential moving average of weights (decay 0.995). Batch size is 4 with 1000 steps per epoch, sampling 1\% of available files each epoch to ensure diversity.

\subsection{Synthetic anomaly evaluation}

For systematic evaluation, we inject synthetic anomalies into validation images not seen during training. Each anomaly is localized within a circular region covering approximately 20\% of the evaluation mask area. Five anomaly types are tested: (1) brightness increase (additive), (2) darkening (subtractive), (3) texture modification (spatially-correlated noise), (4) spectral shift (channel-dependent perturbation), and (5) vegetation change (NIR boost with visible reduction). Anomaly intensity is sampled uniformly from 0.3\% to 25\% of the local dynamic range.

Detection performance is evaluated using ROC-AUC, precision-recall AUC (PR-AUC), and best F1 score. Metrics are computed within the inpainting mask region only, ensuring fair comparison across methods. Results are aggregated over 150 test samples with balanced representation across anomaly types and intensities.

\subsection{Anomaly detection pipeline}

For anomaly detection, we construct the input sequence from the four most recent cloud-free acquisitions preceding the target date. The target frame is masked with a circular region of radius 50 pixels (500~m) at image center. The model predicts reflectance within this region, and anomaly score is computed as the 95th percentile of per-pixel L1 error within an evaluation mask.

The evaluation mask excludes: (1) pixels flagged as cloud, cloud shadow, water, or snow/ice by SCL; (2) pixels in topographic shadow (hillshade $<$ 0.5); (3) pixels in projected cloud shadow based on solar geometry and cloud height estimation.

Topographic shadow is computed from Copernicus DEM-30 data resampled to the Sentinel-2 grid. Solar azimuth and elevation are computed from acquisition time and location using the pysolar library. Cloud shadow projection assumes cloud heights between 500~m and 3000~m, projecting the cloud mask along the solar vector.

\subsection{Baseline methods}

The temporal median predictor computes the pixel-wise median of the three preceding frames and uses this as the prediction for the target frame. Anomaly score is the 95th percentile of per-pixel absolute difference within the evaluation mask.

The RX detector computes local Mahalanobis distance for each pixel in the target frame. The local mean and covariance are estimated from all valid pixels within the evaluation mask. Anomaly score is the 95th percentile of Mahalanobis distance.

All methods use identical evaluation masks for fair comparison.

\subsection{Code and data availability}

Training and inference code are implemented in PyTorch. Sentinel-2 data are freely available through the Copernicus Open Access Hub. The SATLAS pre-trained weights are available at \url{https://github.com/allenai/satlas}.

\section*{Acknowledgements}

B.R.L. acknowledges support from the JSPS Hakubi program. B.R.L and C.H. acknowledge support from NASA NSPIRES (grant 80NSSC22K1282).

\bibliographystyle{naturemag}

\end{document}